# Confidence Calibration for Object Detection and Segmentation

**Fabian Küppers, Anselm Haselhoff, Jan Kronenberger, and Jonas Schneider**

**Abstract** Calibrated confidence estimates obtained from neural networks are crucial, particularly for safety-critical applications such as autonomous driving or medical image diagnosis. However, although the task of confidence calibration has been investigated on classification problems, thorough investigations on object detection and segmentation problems are still missing. Therefore, we focus on the investigation of confidence calibration for object detection and segmentation models in this chapter. We introduce the concept of multivariate confidence calibration that is an extension of well-known calibration methods to the task of object detection and segmentation. This allows for an extended confidence calibration that is also aware of additional features such as bounding box/pixel position and shape information. Furthermore, we extend the expected calibration error (ECE) to measure miscalibration of object detection and segmentation models. We examine several network architectures on MS COCO as well as on Cityscapes and show that especially object detection as well as instance segmentation models are intrinsically miscalibrated given the introduced definition of calibration. Using our proposed calibration methods, we have been able to improve calibration so that it also has a positive impact on the quality of segmentation masks as well.

F. Küppers (✉) · A. Haselhoff · J. Kronenberger
Hochschule Ruhr West, Duisburger Str. 100, 45479 Mülheim a.d. Ruhr, Germany
e-mail: fabian.kueppers@hs-ruhrwest.de

A. Haselhoff
e-mail: anselm.haselhoff@hs-ruhrwest.de

J. Kronenberger
e-mail: jan.kronenberger@hs-ruhrwest.de

J. Schneider
Elektronische Fahrwerksysteme GmbH, Dr.-Ludwig-Kraus-Str. 6,
85080 Gaimersheim, Germany
e-mail: jonas.schneider@efs-auto.com

## 1 Introduction

Common neural networks for object detection must not only determine the position and class of an object but also denote their confidence about the correctness of each detection. This also holds for instance or semantic segmentation models that output a score for each pixel indicating the confidence of object mask membership. A reliable confidence estimate for detected objects is crucial, particularly for safety-critical applications such as autonomous driving, to reliably process detected objects. Another example is medical diagnosis, where, for example, the shape of a brain tumor within an MRI image is of special interest [MWT+20].

Each confidence estimate might be seen as a probability of correctness, reflecting the model's uncertainty about a detection or the pixel mask. During inference, we expect the estimated confidence to match the observed precision for a prediction. For example, given 100 predictions with 80% confidence each, we expect 80 predictions to be correctly predicted [GPSW17, KKSH20]. However, recent work has shown that the confidence estimates of either classification or detection models based on neural networks are *miscalibrated*, i.e., the confidence does not match the observed accuracy in classification [GPSW17] or the observed precision in object detection [KKSH20]. While confidence calibration within the scope of classification has been extensively investigated [NCH15, NC16, GPSW17, KSFF17, KPNK+19], we recently defined the term of calibration for object detection and proposed methods to measure and resolve miscalibration [KKSH20, SKR+21]. In this context, we measured miscalibration w.r.t. the position and scale of detected objects by also including the regression branch of an object detector into a calibration mapping. We have been able to show that modern object detection models also tend to be miscalibrated. On the one hand, this can be mitigated using standard calibration methods. On the other hand, we show that our proposed methods for position- and shape-dependent calibration [KKSH20] are able to further reduce miscalibration.

In this chapter we review our methods for position- and shape-dependent confidence calibration and provide a definition for confidence calibration for the task of *instance/semantic segmentation*. To this end, we extend the definition of the *expected calibration error* (ECE) to measure miscalibration within the scope of instance/semantic segmentation. Furthermore, we adapt the extended calibration methods originally designed for object detection [KKSH20] to enable position-dependent confidence calibration for segmentation tasks as well.

This chapter is structured as follows. In Sect. 2 we review the most important related works regarding confidence calibration. In Sect. 3 we provide the definitions of calibration for the tasks of object detection, instance segmentation, and semantic segmentation. Furthermore, in Sect. 4 we present the extended confidence calibration methods for object detection and segmentation. Extensive experimental evaluations on a variety of architectures and computer vision problems, including object detection and instance/semantic segmentation, are discussed in Sect. 5. Finally, we provide conclusions and discuss our most important findings.

## 2 Related Works

In the past, most research focused on measuring and resolving miscalibration for classification tasks. In this scope, the *expected calibration error* (ECE) [NCH15] is commonly used in conjunction with the Brier score and negative log likelihood (NLL) loss to measure miscalibration. For calculating the ECE, all samples are grouped into equally sized bins by their confidence. Afterward, for each bin the accuracy is calculated and used as an approximation of the accuracy of a single sample in its respective bin. Recently, several extensions such as classwise ECE [KPNK+19], marginal calibration error [KLM19], or adaptive calibration error [NDZ+19] for the evaluation of multi-class problems have also been proposed, where the ECE is evaluated for each class separately. In contrast to previous work, we extend the common ECE definition to the task of object detection [KKSH20] and instance/semantic segmentation. This extension allows for a position-dependent miscalibration evaluation so that it is possible to quantify miscalibration separately for certain image regions. The definition is given in Sect. 3.

Besides measuring miscalibration, it is also possible to resolve a potential miscalibration by using calibration methods that are applied after inference. These post-hoc calibration methods can be divided into binning and scaling methods. Binning methods such as histogram binning [ZE01], isotonic regression [ZE02], Bayesian binning into quantiles (BBQ) [NCH15], or ensemble of near-isotonic regression (ENIR) [NC16] group all samples into several bins by their confidence (similar to the ECE calculation) and perform a mapping from uncalibrated confidence estimates to calibrated ones. In contrast, scaling methods such as logistic calibration (Platt scaling) [Pla99], temperature scaling [GPSW17], beta calibration [KSFF17], or Dirichlet calibration [KPNK+19] scale the network logits before sigmoid/softmax activation to obtain calibrated confidences. The scaling parameters are commonly obtained by logistic regression. Other approaches comprise binwise temperature scaling [JJY+19] or scaling-binning calibrator [KLM19], combining both approaches to further improve calibration performance for classification tasks.

Recently, we proposed an extension of common calibration methods to object detection by also including the bounding box regression branch of an object detector into a calibration mapping [KKSH20]. On the one hand, we extended the histogram binning [ZE01] to perform calibration using a multidimensional binning scheme. On the other hand, we also extended scaling methods to include position information into a calibration mapping. Both approaches are presented in Sect. 4 in more detail.

Unlike classification, the task of instance or semantic segmentation calibration has not yet been addressed by many authors. In the work of [WLK+20], the authors perform online confidence calibration of the classification head of an instance segmentation model and show that this has a significant impact on the mask quality. The authors in [KG20] use a multi-task learning approach for semantic segmentation models to improve model calibration and out-of-distribution detection within the scope of medical image diagnosis. A related approach is proposed by [MWT+20], where the authors train multiple fully-connected networks as an ensemble to obtain well-calibrated semantic segmentation models. However, none of these methods provide an explicit definition of calibration for segmentation models (instance or semantic). This problem is addressed by [DLXS20], where the authors explicitly define semantic segmentation calibration and propose a local temperature scaling for semantic image masks. This approach utilizes the well-known temperature scaling [GPSW17] and assigns a temperature for each mask pixel separately. Furthermore, they use a dedicated convolutional neural network (CNN) to infer the temperature parameters for each image. Our definition of segmentation calibration in Sect. 3 is conform with their definition. Our approach differs from their proposed image-based temperature scaling as we use a single position-dependent calibration mapping to model the probability distributions for all images.

## 3 Calibration Definition and Evaluation

In this section, we review the term of confidence calibration for *classification* tasks [NCH15, GPSW17] and extend this definition to *object detection*, *instance segmentation*, and *semantic segmentation*. Furthermore, we derive the *detection expected calibration error* (D-ECE) to measure miscalibration.

### 3.1 Definitions of Calibration

**Classification:** In a first step, we define the datasets $\mathcal{D}$ of size $|\mathcal{D}| = N$ with indices $i \in \mathcal{I} = \{1, \ldots, N\}$, consisting of images $\mathbf{x} \in \mathbb{I}^{H \times W \times C}$, $\mathbb{I} = [0, 1]$, with height $H$, width $W$, and number of channels $C$. Each image has ground-truth information that consists of the class information $\overline{Y} \in \mathcal{Y} = \{1, \ldots, Y\}$. As an introduction into the task of confidence calibration, we start with the definition of perfect calibration for classification. A classification model $\mathbf{F}_{\text{cls}}$ outputs a label $\hat{Y}$ and a corresponding confidence score $\hat{P}$ indicating its belief about the prediction's correctness. In this case, perfect calibration is defined by [GPSW17]

$$P(\hat{Y} = \overline{Y} | \hat{P} = p) = p \quad \forall p \in [0, 1]. \tag{1}$$

In other words, the accuracy $P(\hat{Y} = \overline{Y} | \hat{P} = p)$ for a certain confidence level $p$ should match the estimated confidence. If we observe a deviation, a model is called *miscalibrated*. In binary classification, we rather consider the *relative frequency* of $\overline{Y} = 1$ instead of the accuracy as the calibration measure. We illustrate this difference using the following example. Consider $N = 100$ images of a dataset $\mathcal{D}$ with binary ground-truth labels $\overline{Y} \in \{0, 1\}$, where 50 images are labeled as $\overline{Y} = 0$ and 50 images with $\overline{Y} = 1$. Furthermore, consider a classification model $\mathbf{F}_{\text{cls}}$ with a sigmoidal output in $[0, 1]$, indicating its confidence for $\overline{Y} = 1$. In our example, this model is able to always predict the correct ground-truth label with confidences $\hat{P} = 0$ and $\hat{P} = 1$ for $\hat{Y} = 0$ and $\hat{Y} = 1$, respectively. Thus, the network is able to achieve an accuracy of 100% but with an average confidence of 50%. Therefore, we consider the relative frequency for $\overline{Y} = 1$ in a binary classification task as the calibration goal that is also 50% in this scenario.

**Object detection:** In the next step, we extend our dataset and model to the task of object detection. In contrast to a classification dataset, an object detection dataset consists of ground-truth annotations $\overline{Y} \in \mathcal{Y} = \{1, \ldots, Y\}$ for each object within an image as well as the ground-truth position and shape information $\overline{\mathbf{R}} \in \mathcal{R} = [0, 1]^A$ ($A$ denotes the size of the normalized box encoding, comprising the center $x$ and $y$ positions $c_x$, $c_y$, as well as width $w$ and height $h$). An object detection model $\mathbf{F}_{\text{det}}$ further outputs a confidence score $\hat{P} \in [0, 1]$, a label $\hat{Y} \in \mathcal{Y}$, and the corresponding position $\hat{\mathbf{R}} \in \mathcal{R}$ for each detection in an image $\mathbf{x}$. Extending the original formulation for confidence calibration within classification tasks [GPSW17], perfect calibration for *object detection* is defined by [KKSH20]

$$P(M = 1 | \hat{P} = p, \hat{Y} = y, \hat{\mathbf{R}} = \mathbf{r}) = p \qquad (2)$$
$$\forall p \in [0, 1], y \in \mathcal{Y}, \mathbf{r} \in \mathcal{R},$$

where $M = 1$ denotes a correctly classified prediction that matches a ground-truth object with a certain intersection-over-union (IoU) score. Commonly, an object detection model is calibrated by means of its precision, as the computation of the accuracy is not possible without knowing all anchors of a model [KKSH20, SKR+21]. Thus, $P(M = 1)$ is a shorthand notation for $p(\hat{Y} = \overline{Y}, \hat{\mathbf{R}} = \overline{\mathbf{R}})$ that expresses the precision for a dedicated IoU threshold.

**Instance segmentation:** At this point, we adapt this idea to define confidence calibration for *instance segmentation*. Consider a dataset with $K$ annotated objects $\mathcal{K}$ over all images. For notation simplicity, we further use $j \in \mathcal{J}_k = \{1, \ldots, H_k \cdot W_k\}$ as the index for pixel $j$ within a bounding box $\overline{\mathbf{R}}_k$ of object $k \in \mathcal{K}$ in the instance segmentation dataset, where $H_k$ and $W_k$ denote the width and height of object $k$, respectively. In addition to object detection, a ground-truth dataset for instance segmentation also consists of pixel-wise mask labels denoted by $\overline{Y}_j \in \mathcal{Y}^* = \{0, 1\}$ for each pixel $j$ in the bounding box $\overline{\mathbf{R}}_k$ of object $k$. Note that we introduce the star superscript (*) here to distinguish between the bounding box label/prediction encoding and the instance segmentation label/prediction encoding. An instance segmentation model $\mathbf{F}_{\text{ins}}$ predicts the membership $\hat{Y}_j \in \mathcal{Y}^*$ for each pixel $j$ in the predicted bounding box $\hat{\mathbf{R}}_k$ to the object mask with a certain confidence $\hat{P}_j \in [0, 1]$. We further denote $\mathbf{R}_j \in \mathcal{R}^* = [0, 1]^{A^*}$ as the position of pixel $j$ *within the bounding box* $\hat{\mathbf{R}}_k$, where $A^*$ denotes the size of the used position encoding of a pixel within its bounding box. In contrast to object detection, it is possible to treat the confidence scores of each pixel within an instance segmentation mask as a binary classification problem. In this case, the confidence $\hat{P}_j$ can be interpreted as the probability of a pixel belonging to the object mask. Therefore, the aim of confidence calibration for *instance segmentation* is that the pixel confidence should match the relative frequency that a pixel is part of the object mask. According to the task of object detection, we further include a position dependency into the definition, for instance, segmentation and obtain

$$P(\hat{Y}_j = \overline{Y}_j | \hat{Y} = y, \hat{P}_j = p, \mathbf{R}_j = \mathbf{r}) = p, \tag{3}$$
$$\forall p \in [0, 1], y \in \mathcal{Y}, \mathbf{r} \in \mathcal{R}^*, j \in \mathcal{J}_k, k \in \mathcal{K}.$$

The former term can be interpreted as the probability that the prediction $\hat{Y}_j$ for a pixel with index $j$ within an object $k$ matches the ground-truth annotation $\overline{Y}_j$ given a certain confidence $p$, a certain pixel position $\mathbf{r}$ within the bounding box, as well as a certain object category $y$ that is predicted by the bounding box head of the instance segmentation model.

**Semantic Segmentation:** Compared to object detection or instance segmentation, a *semantic segmentation* dataset does not hold ground-truth information for individual objects but rather consists of pixel-wise class annotations $\overline{Y}_j \in \mathcal{Y} = \{1, \ldots, Y\}$, $j \in \mathcal{J} = \{1, \ldots, H \cdot W\}$. A semantic segmentation model $\mathbf{F}_{\text{sem}}$ outputs pixel-wise labels $\hat{Y}_j$ and probabilities $\hat{P}_j$ with relative position $\mathbf{R}_j$ *within an image*. Therefore, we can define perfect calibration for *semantic segmentation* as

$$P(\hat{Y}_j = \overline{Y}_j | \hat{P}_j = p, \mathbf{R}_j = \mathbf{r}) = p, \tag{4}$$
$$\forall p \in [0, 1], \mathbf{r} \in \mathcal{R}^*, j \in \mathcal{J},$$

that is related to the calibration definition of classification [GPSW17]. In addition, the confidence score of each pixel must not only reflect its accuracy given a certain confidence level but also at a certain pixel position.

## 3.2 Measuring Miscalibration

We can measure the miscalibration of an *object detection* model as the expected deviation between confidence and observed precision which is also known as the *detection expected calibration error* (D-ECE) [KKSH20]. Let further $\mathbf{s} = (p, y, \mathbf{r})$ denote a single detection with confidence $p$, class $y$, and bounding box $\mathbf{r}$ so that $\mathbf{s} \in \mathcal{S}$, where $\mathcal{S}$ is the aggregated set of the confidence space $[0, 1]$, the set of ground-truth labels $\mathcal{Y}$, and the set of possible bounding box positions $\mathcal{R}$. Since $M$ is a continuous random variable, we need to approximate the D-ECE using the Riemann-Stieltjes integral [GPSW17]

$$\mathbb{E}_{\hat{P}, \hat{Y}, \hat{\mathbf{R}}}[|P(M = 1 | \hat{P} = p, \hat{Y} = y, \hat{\mathbf{R}} = \mathbf{r}) - p|] \tag{5}$$
$$= \int_{\mathcal{S}} \left| P(M = 1 | \hat{P} = p, \hat{Y} = y, \hat{\mathbf{R}} = \mathbf{r}) - p \right| dF_{\hat{P}, \hat{Y}, \hat{\mathbf{R}}}(\mathbf{s}) \tag{6}$$

with $F_{\hat{P}, \hat{Y}, \hat{\mathbf{R}}}(\mathbf{s})$ as the joint cumulative distribution of $\hat{P}$, $\hat{Y}$, and $\hat{\mathbf{R}}$. Let $\mathbf{a} = (a_p, a_y, a_{r_1}, \ldots, a_{r_A})^\mathsf{T} \in \mathcal{S}$ and $\mathbf{b} = (b_p, b_y, b_{r_1}, \ldots, b_{r_A})^\mathsf{T} \in \mathcal{S}$ denote interval boundaries for the confidence $a_p, b_p \in [0, 1]$, the estimated labels $a_y, b_y \in \mathcal{Y}$ and each quantity of the bounding box encoding $a_{r_1}, b_{r_1}, \ldots a_{r_A}, b_{r_A} \in [0, 1]$ defined on the joint cumulative $F_{\hat{P},\hat{Y},\hat{R}}(s)$ so that[1] $F_{\hat{P},\hat{Y},\hat{R}}(\mathbf{b}) - F_{\hat{P},\hat{Y},\hat{R}}(\mathbf{a}) = P(\hat{P} \in [a_p, b_p], \hat{Y} \in [a_y, b_y], \hat{R}_1 \in [a_{r_1}, b_{r_1}], \ldots, \hat{R}_A \in [a_{r_A}, b_{r_A}])$. The integral in (6) is now approximated by

$$\sum_{m \in \mathcal{B}} P(\mathbf{s}_m) \cdot \left| P(M = 1 | \hat{P} = p_m, \hat{Y} = y_m, \hat{\mathbf{R}} = \mathbf{r}_m) - p_m \right|, \quad (7)$$

with $B$ as the number of equidistant bins used for integral approximation so that $\mathcal{B} = \{1, \ldots, B\}$, and $p_m \in [0, 1]$, $y_m \in \mathcal{Y}$ and $\mathbf{r}_m \in \mathcal{R}$ being the respective bin entities for average confidence, the current label, and the average (unpacked) bounding box scores within bin $m \in \mathcal{B}$, respectively. Let $N_m$ denote the amount of samples within a single bin $m$. For large datasets $|\mathcal{D}| = N$, the probability $P(M = 1 | \hat{P} = p_m, \hat{Y} = y_m, \hat{\mathbf{R}} = \mathbf{r}_m)$ is approximated by the average precision within a single bin $m$, whereas $p_m$ is approximated by the average confidence, so that the D-ECE can finally be computed by

$$\sum_{m \in \mathcal{B}} \frac{1}{N_m} \left| \text{prec}(m) - \text{conf}(m) \right|, \quad (8)$$

where $\text{prec}(m)$ and $\text{conf}(m)$ denote the precision and average confidence within bin $m$, respectively.

Similarly, we can extend the D-ECE to *instance segmentation* by

$$\mathbb{E}_{\hat{Y}, \hat{P}_j, \mathbf{R}_j}[|P(\hat{Y}_j = \overline{Y}_j | \hat{Y} = y, \hat{P}_j = p, \mathbf{R}_j = \mathbf{r}) - p|] \quad (9)$$

$$\approx \sum_{m \in \mathcal{B}} \frac{1}{N_m} \left| \text{freq}(m) - \text{conf}(m) \right|, \quad (10)$$

with a binning scheme over all pixels and freq(m) as the average frequency within each bin. In this case, each pixel is treated as a separate prediction and binned by its confidence, label class, and relative position. In the same way, the D-ECE for *semantic segmentation* is approximated by

$$\mathbb{E}_{\hat{P}_j, \mathbf{R}_j}[|P(\hat{Y}_j = \overline{Y}_j | \hat{P}_j = p, \mathbf{R}_j = \mathbf{r}^*) - p|] \quad (11)$$

$$\approx \sum_{m \in \mathcal{B}} \frac{1}{N_m} \left| \text{acc}(m) - \text{conf}(m) \right|, \quad (12)$$

with accuracy $\text{acc}(m)$ within each bin $m$.

---

[1] For discrete random variables (e.g., $Y$), we use a continuous density function of the form $p_Y(y) = \sum_{y^* \in \mathcal{Y}} P(y^*) \delta(y - y^*)$ with $\delta(x)$ as the Dirac delta function.

For the calibration evaluation of object detection models, we can use the relative position $c_x$, $c_y$, and the shape $h$, $w$, of the bounding boxes [KKSH20]. For segmentation, we consider the relative position $x$, $y$, of each pixel within a bounding box (instance segmentation) or within the image (semantic segmentation), as well as its distance $d$ to the next segment boundary, as we expect a higher uncertainty in the peripheral areas of a segmentation mask. We use these definitions to evaluate different models in Sect. 5.

## 4 Position-Dependent Confidence Calibration

For post-hoc calibration, we distinguish between binning and scaling methods. According to the approximation in (7), binning methods such as histogram binning [ZE01] divide all samples into several bins by their confidence and measure the average accuracy/precision within each bin. In contrast, scaling methods rescale the logits of a neural network before sigmoid/softmax activation to calibrate a network's output. In this section, we extend standard calibration methods so that they are capable of dealing with additional information such as position and/or shape. These extended methods can be used for confidence calibration of *object detection*, *instance segmentation*, and *semantic segmentation* tasks. We illustrate the difference between standard calibration and position-dependent calibration using an artificially created dataset in Fig. 1. This dataset consists of points with a confidence score and a binary ground-truth information in {0, 1}, so that we are able to compute the frequency of the points across the whole image. We observe that standard calibration only shifts the average confidence to fit the average precision/accuracy. This leads to an

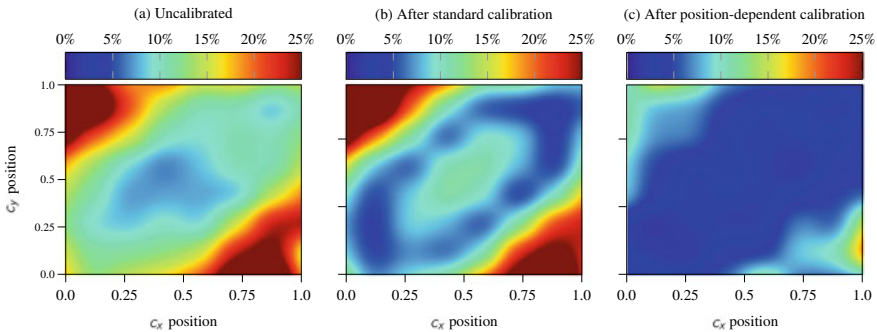

**Fig. 1 a** Consider an artificial dataset with low calibration error in the center and increasing miscalibration toward the image boundaries. **b** A common logistic calibration model only shifts the average confidence to match the accuracy. This leads to a better global ECE score but also degrades the miscalibration in the center of the image. **c** In contrast, position-dependent calibration is capable of possible dependencies and leads to an overall improvement of calibration (example taken from previous work [KKSH20]) increased calibration error in the center of the image. In contrast, position-dependent calibration as defined in this section is able to capture correlations between position information and calibration error and reduces the D-ECE across the whole image.

## 4.1 Histogram Binning

Given a binning scheme with $B$ different bins so that $\mathcal{B} = \{1, \ldots, B\}$ with the according bin boundaries $0 = a_1 \leq a_2 \leq \cdots \leq a_{B+1} = 1$ and the corresponding calibrated estimate $\theta_m$ within each bin, the objective for histogram binning to estimate $\Theta = \{\theta_m | m \in \mathcal{B}\}$ is the minimization

$$\min_{\Theta} \sum_{i \in \mathcal{I}} \sum_{m \in \mathcal{B}} \mathbb{1}(a_m \leq \hat{p}_i < a_{m+1}) \cdot (\theta_m - \bar{y}_i)^2, \tag{13}$$

with $\mathbb{1}(\cdot)$ as the indicator function, yielding a 1 if its argument is true and a 0 if it is false [ZE01, GPSW17]. This objective converges to the fraction of positive samples within each bin under consideration. However, within the scope of object detection or segmentation, we also want to measure the accuracy/precision w.r.t. position and shape information. As an extension to the standard histogram binning [ZE01], we therefore propose a multidimensional binning scheme that divides all samples into several bins by their confidence **and** by all additional information such as position and shape [KKSH20]. We further denote $\hat{\mathbf{S}} = (\hat{P}, \hat{\mathbf{R}})$ of size $Q = A + 1$ as the input vector to a calibration function consisting of the confidence and the bounding box encoding, so that $\mathcal{Q} = \{1, \ldots, Q\}$. In a multidimensional histogram binning, we indicate the number of bins as $\mathbf{B} = (B_1, \ldots, B_Q)$ so that $\mathcal{B}^* = \{\mathcal{B}_q = \{1, \ldots, B_q\} | q \in \mathcal{Q}\}$ with bin boundaries $\{0 = a_{1,q} \leq a_{2,q} \leq \cdots \leq a_{B_q+1,q} = 1 | q \in \mathcal{Q}\}$. For each bin combination $m_1 \in \mathcal{B}_1, \ldots, m_Q \in \mathcal{B}_Q$, we have a dedicated calibration parameter $\theta_{m_1,\ldots,m_Q} \in \mathbb{R}$ so that the calibration parameters $\Theta^*$, with $|\Theta^*| = \prod_{q \in \mathcal{Q}} B_q$, are given as $\Theta^* = \{\theta_{m_1,\ldots,m_Q} | m_1 \in \mathcal{B}_1, \ldots, m_Q \in \mathcal{B}_Q\}$. This results in an objective function given by

$$\min_{\Theta^*} \sum_{i \in \mathcal{I}} J(\hat{\mathbf{s}}_i), \tag{14}$$

where

$$J(\hat{\mathbf{s}}_i) = \begin{cases} (\theta_{m_1,\ldots,m_Q} - \bar{y}_i)^2, & \text{if } (a_{m,q} \leq \hat{s}_{i,q} < a_{m,q}) \quad \forall q \in \mathcal{Q}, \forall m \in \mathcal{B}_q \\ 0, & \text{otherwise}, \end{cases} \tag{15}$$

which again converges to the fraction of positive samples within each bin. The term $J(\hat{\mathbf{s}}_i)$ simply denotes that the loss is only applied if a sample $\hat{\mathbf{s}}_i$ falls in a certain bin combination. A drawback of using this method is that the number of bins is given by $\prod_{q \in \mathcal{Q}} B_q$, which grows exponentially as the number of dimensions $Q$ grows.

## *4.2 Scaling Methods*

As opposed to binning methods like histogram binning, scaling methods perform a rescaling of the logits before sigmoid/softmax activation to obtain calibrated confidences. We can distinguish between logistic calibration (Platt scaling) [Pla99] and beta calibration [KSFF17].

**Logistic calibration:** According to the well-known logistic calibration (Platt scaling) [Pla99], the calibration parameters are commonly obtained using logistic regression. For a binary logistic regression model, we assume normally distributed scores for the positive and negative class, so that $p(p|+) \sim \mathcal{N}(p; \mu_+, \sigma^2)$ and $p(p|-) \sim \mathcal{N}(p; \mu_-, \sigma^2)$. The mean values for the two classes are given by $\mu_+$, $\mu_-$ and the variance $\sigma^2$ is equal for all classes. First, we follow the derivation of logistic calibration introduced by [Pla99, KSFF17] using the likelihood ratio

$$LR(p) = \frac{p(p|+)}{p(p|-)} = \exp\left(\frac{1}{2\sigma^2}[-(p-\mu_+)^2 + (p-\mu_-)^2]\right) \quad (16)$$

$$= \exp\left(\frac{1}{2\sigma^2}[2p(\mu_+ - \mu_-) - (\mu_+^2 - \mu_-^2)]\right) \quad (17)$$

$$= \exp\left(\frac{\mu_+ - \mu_-}{2\sigma^2}[p - (\mu_+ + \mu_-)]\right) \quad (18)$$

$$= \exp(\gamma(p - \eta)), \quad (19)$$

where $\gamma = \frac{1}{2\sigma^2}(\mu_+ - \mu_-)$ and $\eta = \mu_+ + \mu_-$. Assuming a uniform prior over the positive and negative classes, the likelihood ratio equals the posterior odds. Hence, a calibrated probability is derived by

$$P(+|\hat{p}) = \frac{1}{1 + LR(\hat{p})^{-1}} = \frac{1}{1 + \exp(-\gamma(\hat{p} - \eta))}, \quad (20)$$

which recovers the logistic function.

Recently, [KKSH20] used this formulation to derive a position-dependent confidence calibration by using multivariate Gaussians for the positive and negative classes. Introducing the concept of multivariate confidence calibration [KKSH20], we can derive a likelihood ratio for position-dependent logistic calibration by

$$LR_{LC}(\hat{\mathbf{s}}) = \exp\left(\frac{1}{2}\left[(\hat{\mathbf{s}}_-^\mathsf{T} \mathbf{\Sigma}_-^{-1} \hat{\mathbf{s}}_-) - (\hat{\mathbf{s}}_+^\mathsf{T} \mathbf{\Sigma}_+^{-1} \hat{\mathbf{s}}_+)\right] + c\right), \quad c = \log\frac{|\mathbf{\Sigma}_-|}{|\mathbf{\Sigma}_+|}, \quad (21)$$

where $\hat{\mathbf{s}}_+ = \hat{\mathbf{s}} - \boldsymbol{\mu}_+$ and $\hat{\mathbf{s}}_- = \hat{\mathbf{s}} - \boldsymbol{\mu}_-$ using $\boldsymbol{\mu}_+, \boldsymbol{\mu}_- \in \mathbb{R}^Q$ as the mean vectors and $\mathbf{\Sigma}_+, \mathbf{\Sigma}_- \in \mathbb{R}^{Q \times Q}$ as the covariance matrices for the positive and negative classes, respectively.

**Beta calibration:** Similarly, we can also extend the beta calibration method [KSFF17] to a multivariate calibration scheme. However, we need a special form of a multivariate beta distribution as the Dirichlet distribution is only defined over a $Q$-simplex and thus is not suitable for this kind of calibration. Therefore, we use a multivariate beta distribution proposed by [LN82] which is defined as

$$p(\hat{\mathbf{s}}|\boldsymbol{\alpha}) = \frac{1}{B(\boldsymbol{\alpha})} \prod_{q \in Q} \left[ \lambda_q^{\alpha_q} (\hat{s}_q^*)^{\alpha_q - 1} \left( \frac{\hat{s}_q^*}{\hat{s}_q} \right)^2 \right] \left[ 1 + \sum_{q \in Q} \lambda_q \hat{s}_q^* \right]^{-\sum_{q \in Q_0} \alpha_q}, \quad (22)$$

with $Q_0 = \{0, \ldots, Q\}$ and the shape parameters $\boldsymbol{\alpha} = (\alpha_0, \ldots, \alpha_q)^\mathsf{T}$, $\boldsymbol{\beta} = (\beta_0, \ldots, \beta_q)^\mathsf{T}$ that are restricted to $\alpha_q, \beta_q > 0$. Furthermore, we denote $\lambda_q = \frac{\beta_q}{\beta_0}$ and $\hat{s}_q^* = \frac{\hat{s}_q}{1-\hat{s}_q}$. In this context, $B(\boldsymbol{\alpha})$ denotes the multivariate beta function. However, this kind of beta distribution is only able to capture positive correlations [LN82]. Nevertheless, it is possible to derive a likelihood ratio given by

$$LR_{BC}(\hat{\mathbf{s}}) = \exp\left( \sum_{q \in Q} \left[ \alpha_q^+ \log(\lambda_q^+) - \alpha_q^- \log(\lambda_q^-) + (\alpha_q^+ - \alpha_q^-) \log(\hat{s}_q^*) \right] + \quad (23) \right.$$
$$\left. \sum_{q \in Q_0} \left[ \alpha_q^- \log\left( \sum_{j \in Q} \lambda_j^- \hat{s}_j^* \right) - \alpha_q^+ \log\left( \sum_{j \in Q} \lambda_j^+ \hat{s}_j^* \right) \right] + c \right),$$

with $\boldsymbol{\alpha}^+, \boldsymbol{\alpha}^-$ and $\boldsymbol{\lambda}^+, \boldsymbol{\lambda}^-$ as the shape parameters for the multivariate beta distribution in (22) and for the positive and negative class, respectively, so that $c = \log\left( \frac{B(\boldsymbol{\alpha}^-)}{B(\boldsymbol{\alpha}^+)} \right)$. We investigate the effect of position-dependent calibration in the next section using these methods.

## 5 Experimental Evaluation and Discussion

In this section, we evaluate our proposed calibration methods for the tasks of object detection, instance segmentation, and semantic segmentation using pretrained neural networks. For calibration evaluation, we use the MS COCO validation dataset [LMB+14] consisting of 5,000 images with 80 different object classes for object detection and instance segmentation. For semantic segmentation, we use the panoptic segmentation annotations consisting of 171 different object and stuff categories in total. Our investigations are limited to the validation dataset since the training set has already been used for network training and no ground-truth annotations are available for the respective test dataset. Thus, we use a 50%/50% random split and use the first set for calibration training, while the second set is used for the evaluation of the calibration methods. Furthermore, we also utilize the Cityscapes validation dataset [COR+16] consisting of 500 images with 19 different classes that are used for model training. The Munster and Lindau images are used for calibration training, whereas the Frankfurt images are held back for calibration evaluation.

## 5.1 Object Detection

We evaluate our proposed calibration methods *histogram binning* (HB) (14), *logistic calibration* (LC) (21) and *beta calibration* (BC) (23) for the task of object detection using a pretrained `Faster R-CNN X101-FPN` [RHGS15, WKM+19] as well as a pretrained `RetinaNet` [LGG+17, WKM+19] on the MS COCO validation set. For Cityscapes, we use a pretrained `Mask R-CNN R50-FPN` [HGDG17, WKM+19], where only the predicted bounding box information is used for calibration. For calibration evaluation, we use the proposed D-ECE with the same subsets of data that have been used for calibration training. Thus, we use a binning scheme of $B = 20$ using the confidence $\hat{p}$ only. In contrast, we use $B_q = 5$ for $Q = 5$ when all auxiliary information is used. Each bin with less than 8 samples is neglected to increase D-ECE robustness. We further measure the *Brier score* (BS) and the *negative log likelihood* (NLL) of each model to evaluate its calibration properties. It is of special interest to assess if calibration has an influence to the precision/recall. Thus, we also denote the *area under precision/recall curve* (AUPRC) for each model. As opposed to previous examinations [KKSH20], we evaluate calibration using *all* available classes within each dataset. Each of these scores is measured for each class separately. In our experiments, we denote weighted average scores that are weighted by the amount of samples for each class. We perform calibration using IoU scores of 0.50 and 0.75, respectively. Furthermore, only bounding boxes with a score over 0.3 are used for calibration to reduce the amount of non-informative predictions. We give a qualitative example Fig. 2 that illustrates the effect of confidence calibration in the scope of object detection.

In our experiments, we first apply the standard calibration methods (Table 1) and compare the results with the baseline miscalibration of the detection model. We can already observe a default miscalibration of each examined network. This miscalibration is reduced by standard calibration where the scaling methods offer the best performance compared to the histogram binning. In a next step, we apply our box-sensitive calibration that includes the confidence $\hat{p}$, position $c_x$, $c_y$ and shape $w$ and $h$ into a calibration mapping (Table 2). Similar to the confidence-only case in Table 1, the scaling methods consistently outperform baseline and histogram binning calibration.

In each calibration case, we observe a miscalibration of the base network that is alleviated using our proposed calibration methods. By examining the reliability diagram (Fig. 3), we observe that the networks are consistently miscalibrated for all confidence levels. This can be alleviated either by standard calibration or by position-dependent calibration. By examining the position-dependence of the

Confidence Calibration for Object Detection and Segmentation 237

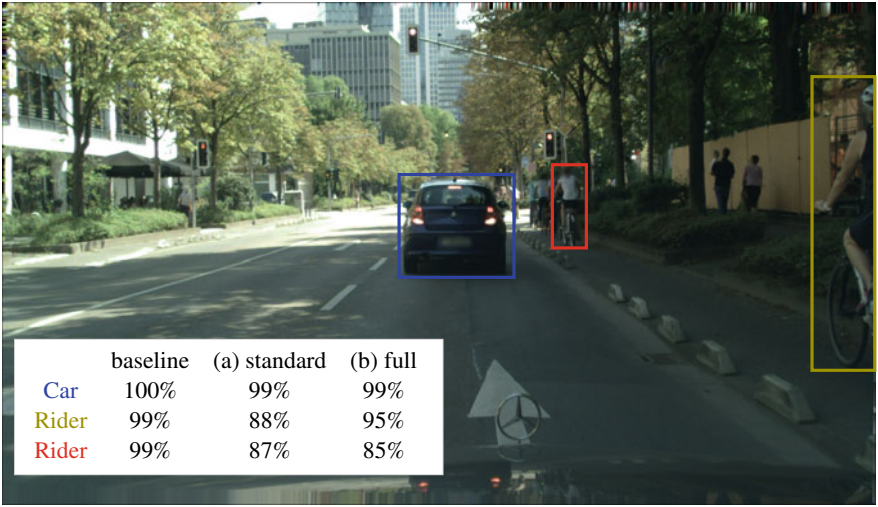

**Fig. 2** Qualitative example of position-dependent confidence calibration on a Cityscapes image with detections obtained by a `Mask-RCNN`. First, standard calibration is performed using similar confidence estimates $\hat{p}$ without any position information. This results in rescaled but still similar calibrated confidence estimates (note that we have distinct calibration models for each class). Second, we can observe the effect of position-dependent calibration using the full position information (position and shape) as the two riders with similar confidence are rescaled differently. When using all available information, we can observe a significant difference in calibration since both, position and shape, have an effect to the calibration

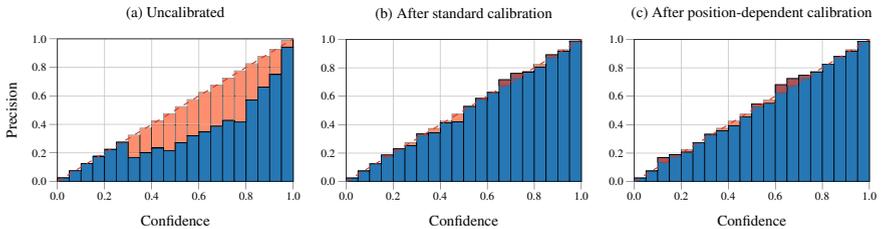

**Fig. 3** Reliability diagram (object detection) for a `Mask-RCNN` class *pedestrian* inferred on Cityscapes Frankfurt images measured using the confidence only. The blue bar denotes the precision of samples that fall into this bin, with the gap to perfect calibration (diagonal) highlighted in red. **a** The `Mask-RCNN` model is consistently overconfident for each confidence level. **b** This can be alleviated by standard logistic calibration **c** as well as by position-dependent logistic calibration

**Table 1** Calibration results for object detection using the confidence $\hat{p}$ only. The best scores are highlighted in bold. All calibration methods are able to improve miscalibration. The best performance is achieved using the scaling methods. Unlike histogram binning, the scaling methods are monotonically increasing and thus do not affect the AUPRC score

| Network | Dataset | IoU | Cal. method | D-ECE (%) | Brier | NLL | AUPRC |
|---|---|---|---|---|---|---|---|
| Faster R-CNN | COCO | 0.50 | Baseline | 17.839 | 0.206 | 0.622 | 0.833 |
| | | | HB | 5.622 | 0.166 | 0.598 | 0.781 |
| | | | LC | 4.449 | **0.158** | **0.478** | 0.833 |
| | | | BC | **4.441** | **0.158** | 0.482 | 0.833 |
| | | 0.75 | Baseline | 29.721 | 0.272 | 0.861 | 0.767 |
| | | | HB | 5.173 | 0.158 | 0.594 | 0.676 |
| | | | LC | **4.387** | **0.147** | **0.461** | 0.766 |
| | | | BC | 4.540 | **0.147** | 0.463 | 0.766 |
| RetinaNet | COCO | 0.50 | Baseline | 10.061 | 0.171 | 0.514 | 0.831 |
| | | | HB | 5.332 | 0.166 | 0.579 | 0.795 |
| | | | LC | 4.464 | **0.161** | **0.486** | 0.831 |
| | | | BC | **4.443** | **0.161** | 0.487 | 0.831 |
| | | 0.75 | Baseline | 17.543 | 0.181 | 0.544 | 0.765 |
| | | | HB | 5.773 | 0.147 | 0.551 | 0.721 |
| | | | LC | 4.395 | **0.142** | **0.447** | 0.765 |
| | | | BC | **4.361** | **0.142** | **0.447** | 0.765 |
| Mask R-CNN | Cityscapes | 0.50 | Baseline | 10.822 | 0.146 | 0.500 | 0.951 |
| | | | HB | 3.516 | 0.133 | 0.491 | 0.902 |
| | | | LC | **3.305** | **0.125** | **0.380** | 0.951 |
| | | | BC | 3.306 | **0.125** | 0.381 | 0.951 |
| | | 0.75 | Baseline | 29.530 | 0.271 | 1.063 | 0.893 |
| | | | HB | **3.718** | 0.161 | 0.547 | 0.755 |
| | | | LC | 4.399 | **0.136** | **0.425** | 0.893 |
| | | | BC | 4.299 | **0.136** | 0.426 | 0.893 |

miscalibration (Figs. 4 and 5), a considerable increase of the calibration error toward the image boundaries can be observed. This calibration error is already well mitigated by standard calibration methods and can be further improved by position-dependent calibration. Also note that the AUPRC is not affected by standard scaling methods (logistic/beta calibration) as these methods perform a monotonically increasing mapping of the confidence estimates and thus do not affect the order of the samples. However, this is not the case with histogram binning which may lead to a significant drop of the AUPRC. Furthermore, even the position-dependent scaling methods cannot guarantee a monotonically increasing mapping. However, compared to the improvement of the calibration, the impact on the AUPRC is marginal. Therefore, we conclude that our calibration methods are a valuable contribution especially for safety-critical systems, since they lead to statistically better calibrated confidences.

**Table 2** Calibration results for object detection using all available information $\hat{p}$, $c_x$, $c_y$, $w$ and $h$. Similar to the confidence-only case in Table 1, the scaling methods consistently outperform baseline and histogram binning calibration. However, in the position-dependent case, the calibration mapping is not monotonically increasing. This has an effect to the AUPRC scores, but is marginal compared to the improvement in calibration

| Network | Dataset | IoU | Cal. method | D-ECE [%] | Brier | NLL | AUPRC |
|---|---|---|---|---|---|---|---|
| Faster R-CNN | COCO | 0.50 | Baseline | 7.519 | 0.206 | 0.622 | **0.833** |
| | | | HB | 3.891 | 0.206 | 1.076 | 0.701 |
| | | | LC | **3.100** | 0.175 | 0.627 | 0.807 |
| | | | BC | 3.240 | **0.168** | **0.508** | 0.807 |
| | | 0.75 | Baseline | 13.277 | 0.272 | 0.861 | **0.767** |
| | | | HB | 3.848 | 0.200 | 1.037 | 0.587 |
| | | | LC | **2.965** | 0.162 | 0.587 | 0.730 |
| | | | BC | 3.217 | **0.159** | **0.492** | 0.719 |
| RetinaNet | COCO | 0.50 | Baseline | 4.303 | **0.171** | 0.514 | **0.831** |
| | | | HB | 3.369 | 0.199 | 1.100 | 0.727 |
| | | | LC | **3.124** | 0.177 | 0.615 | 0.804 |
| | | | BC | 3.491 | 0.175 | 0.521 | 0.800 |
| | | 0.75 | Baseline | 6.724 | 0.181 | 0.544 | **0.765** |
| | | | HB | 3.241 | 0.176 | 1.026 | 0.640 |
| | | | LC | **2.872** | 0.157 | 0.595 | 0.732 |
| | | | BC | 3.205 | **0.153** | **0.479** | 0.725 |
| Mask R-CNN | Cityscapes | 0.50 | Baseline | 10.168 | 0.146 | 0.500 | **0.951** |
| | | | HB | 5.241 | 0.151 | 0.535 | 0.854 |
| | | | LC | **4.551** | **0.134** | 0.440 | 0.946 |
| | | | BC | 6.117 | **0.134** | **0.413** | 0.925 |
| | | 0.75 | Baseline | 27.066 | 0.271 | 1.063 | 0.893 |
| | | | HB | 8.062 | 0.195 | 0.607 | 0.681 |
| | | | LC | **6.267** | **0.140** | **0.464** | **0.894** |
| | | | BC | 9.427 | 0.160 | 0.491 | 0.833 |

## 5.2 Instance Segmentation

After object detection, we investigate the calibration properties of instance segmentation models. According to the definition of calibration for segmentation models in (9), we can use each pixel within a segmentation mask as a separate prediction.

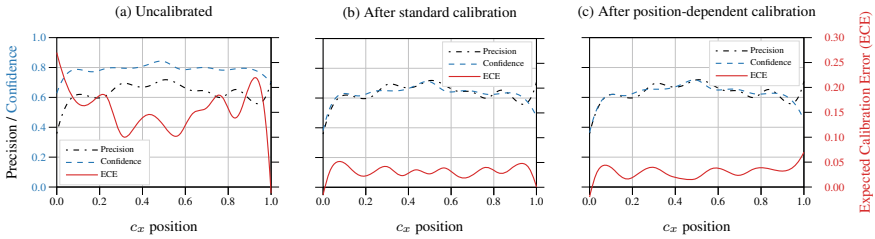

**Fig. 4** Reliability diagram (object detection) for a `Mask-RCNN` class *pedestrian* inferred on Cityscapes Frankfurt images measured by the confidence and the relative $c_x$ position of each bounding box. **a** We observe an increasing calibration error toward the boundaries of the images. **b** This can be alleviated by standard logistic calibration **c** as well as by position-dependent logistic calibration

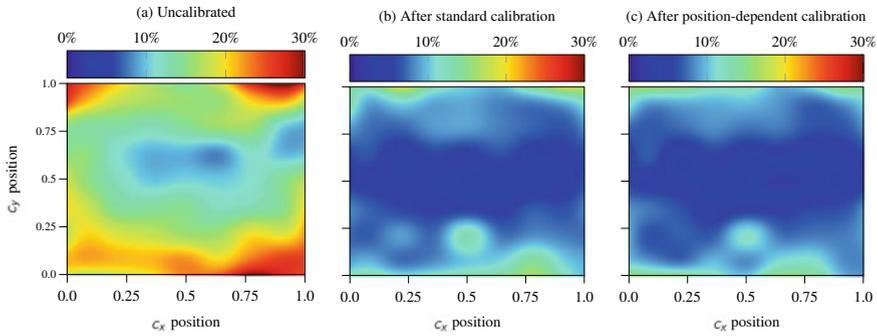

**Fig. 5** Reliability diagram (object detection) for a `Mask-RCNN` class *pedestrian* measured using all available information. This diagram shows the D-ECE [%] for all image regions. **a** We can observe a default miscalibration that increases toward the image boundaries. **b** Standard logistic calibration already achieves good calibration results, **c** that can be further improved by position-dependent calibration

This alleviates the problem of limited data availability and allows for a more robust calibration evaluation. Recently, Kumar et al. [KLM19] started a discussion about the sample-efficiency of binning and scaling methods. The authors show that binning methods yield a more robust calibration mapping for large datasets but also tend to overfit for a small database. We can confirm this observation as histogram binning provides poor calibration performance in our examinations on object detection calibration, particularly for classes with fewer samples. In contrast, scaling methods are more sample-efficient but also more inaccurate [KLM19]. Furthermore, scaling methods are computationally more expensive as they require an iterative update of the calibration parameters over the complete dataset, whereas a binning of samples comes at low computational costs, especially for large datasets. Therefore, our examinations are focused on the calibration performance of a (multivariate) histogram binning using 15 bins for each dimension. For inference, we use pretrained `Mask R-CNN` [HGDG17, WKM+19] as well as pretrained `PointRend` [KWHG20] models to obtain predictions of instance segmentation masks for both datasets. As within object detection evaluation, all objects with a bounding box score below 0.3 are neglected. We perform standard calibration using the confidence only, as well as a position-dependent calibration including the *x* and *y* position of each pixel. The pixel position is scaled by the mask's bounding box size to get position information in the [0, 1] interval. Furthermore, we also include the distance of each pixel to the nearest mask segment boundary as a feature in a calibration mapping since we expect a higher uncertainty especially at the segment boundaries. The distance is normalized by the bounding box's diagonal to obtain distance values in [0, 1].

Since many data samples are available, we measure calibration using a D-ECE with 15 bins neglecting each bin with less than 8 samples. We further assess the Brier score as well as the NLL loss as complementary metrics. The task of instance segmentation is related to object detection. In a first step, the network also needs to infer a bounding box for each detected object. Similar to the calibration evaluation for object detection, we further use IoU scores of 0.50 and 0.75 to specify whether a prediction has matched a ground-truth object or not. In contrast to object detection, the IoU score within instance segmentation is computed by the overlap of the inferred segmentation mask and the according ground-truth object. Using this definition, we can compute the AUPRC to evaluate the average quality of the object detection branch as well as the according segmentation masks. All results for instance segmentation calibration for IoU of 0.50 and 0.75 are shown in Tables 3 and 4, respectively. These tables compare standard calibration (subset: confidence only) with position-dependent calibration (subset: full). We observe a significant miscalibration of the networks by default. Using standard calibration or our calibration methods, it is possible to improve the calibration score D-ECE, Brier, and NLL for each case. The miscalibration is successfully corrected by histogram binning using either the standard binning or the position-dependent binning. This is also underlined by inspecting the reliability diagrams for the confidence-only case (Fig. 6). Furthermore, we can also observe miscalibration that is dependent on the relative *x* and *y* pixel position within a mask (Figs. 7 and 8). We can observe that standard histogram binning calibration already achieves good calibration results but also offers a weak dependency on the pixel position as well. Although the position-dependent calibration only achieves a minor improvement in calibration compared to the standard calibration case, it shows an equal calibration performance across the whole image

In contrast to object detection, we observe a slightly increased miscalibration toward the center of a mask. As it can be seen in Fig. 9, most pixels belonging to an object mask are also located in the center. This underlines the need for a position-dependent calibration. Interestingly, although position-dependent calibration does not seem to offer better Brier or NLL scores compared to the confidence-only case, it significantly improves the mask quality as we can observe higher AUPRC scores for position-dependent calibration. We further provide a qualitative example that illustrates the effect of confidence calibration in Fig. 9. In this example, we can see the difference between standard calibration and position-dependent calibration. In the former case, the mask scores are only rescaled by their confidence which might lead to a better calibration but sometimes also to unwanted losses of mask segments

**Table 3** Calibration results for instance segmentation @ IoU=0.50. The best D-ECE scores are underlined *for each subset separately*, since it is not convenient to compare D-ECE scores with different subsets to each other [KKSH20]. Furthermore, the best Brier, NLL, and AUPRC scores are highlighted in bold. These scores are only calculated using the confidence information and thus can be compared to each other. The histogram binning calibration consistently improves miscalibration. Furthermore, we find that although position dependence does not improve the Brier and NLL scores as in the standard case, it leads to a significant improvement in the mIoU score

| Network | Dataset | Cal. method | Subset: confidence only | | | | Subset: full | | | |
|---|---|---|---|---|---|---|---|---|---|---|
| | | | D-ECE [%] | Brier | NLL | AUPRC | D-ECE [%] | Brier | NLL | AUPRC |
| Mask R-CNN | CS | Baseline | 7.088 | 0.110 | 0.432 | 0.724 | 11.205 | 0.110 | 0.432 | 0.724 |
| | | HB | 5.661 | **0.099** | **0.320** | 0.723 | 10.119 | 0.117 | 0.530 | **0.787** |
| | COCO | Baseline | 21.983 | 0.222 | 0.940 | 0.663 | 23.409 | 0.222 | 0.940 | 0.663 |
| | | HB | 6.420 | **0.150** | **0.442** | 0.662 | 13.640 | 0.171 | 0.776 | **0.760** |
| PointRend | CS | Baseline | 12.893 | 0.160 | 0.785 | 0.709 | 20.858 | 0.160 | 0.785 | 0.709 |
| | | HB | 2.706 | **0.105** | **0.326** | 0.698 | 19.021 | 0.190 | 1.299 | **0.758** |
| | COCO | Baseline | 22.284 | 0.222 | 0.946 | 0.672 | 23.973 | 0.222 | 0.946 | 0.672 |
| | | HB | 6.348 | **0.144** | **0.428** | 0.664 | 16.060 | 0.180 | 1.005 | **0.751** |

**Table 4** Calibration results for instance segmentation @ IoU=0.75. The best scores are highlighted. We observe the same behavior in calibration for all models as for the IoU=0.50 case shown in Table 3

| Network | Dataset | Cal. method | Subset: confidence only | | | | Subset: full | | | |
|---|---|---|---|---|---|---|---|---|---|---|
| | | | D-ECE [%] | Brier | NLL | AUPRC | D-ECE [%] | Brier | NLL | AUPRC |
| Mask R-CNN | CS | Baseline | 12.862 | 0.147 | 0.622 | 0.375 | 14.470 | 0.147 | 0.622 | 0.375 |
| | | HB | 5.895 | **0.108** | **0.340** | 0.375 | 10.273 | 0.120 | 0.523 | **0.479** |
| | COCO | Baseline | 26.559 | 0.250 | 1.070 | 0.237 | 27.219 | 0.250 | 1.070 | 0.237 |
| | | HB | 6.006 | **0.144** | **0.423** | 0.235 | 12.888 | 0.165 | 0.720 | **0.425** |
| PointRend | CS | Baseline | 18.668 | 0.192 | 0.929 | 0.347 | 25.386 | 0.192 | 0.929 | 0.347 |
| | | HB | 3.899 | **0.115** | **0.349** | 0.344 | 18.415 | 0.191 | 1.247 | **0.486** |
| | COCO | Baseline | 26.643 | 0.248 | 1.060 | 0.258 | 27.377 | 0.248 | 1.060 | 0.258 |
| | | HB | 6.708 | **0.138** | **0.411** | 0.238 | 15.344 | 0.173 | 0.936 | **0.388** |

(especially small objects in the background). In contrast, position-dependent calibration is capable of possible correlations between confidence and pixel position or the size of an object. This leads to improved estimates of the mask confidences even for smaller objects. Therefore, we conclude that multidimensional confidence calibration has a positive influence on the calibration properties as well as on the quality of the object masks.

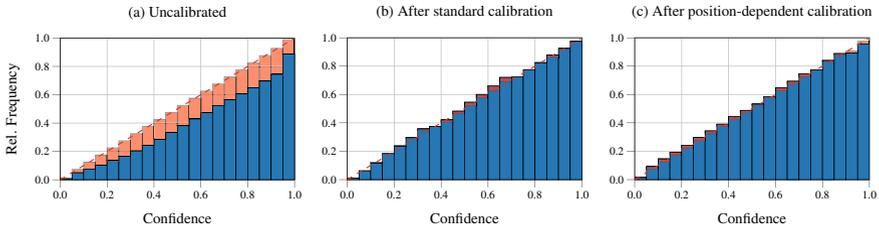

**Fig. 6** Reliability diagram (instance segmentation) for a `Mask R-CNN` class *pedestrian* inferred on MS COCO validation images measured using the confidence only. The blue bar denotes the frequency of samples that fall into this bin, with the gap to perfect calibration (diagonal) highlighted in red. **a** The `Mask R-CNN` model is consistently overconfident for each confidence level. **b** This can be alleviated by standard histogram binning calibration **c** as well as by position-dependent logistic calibration

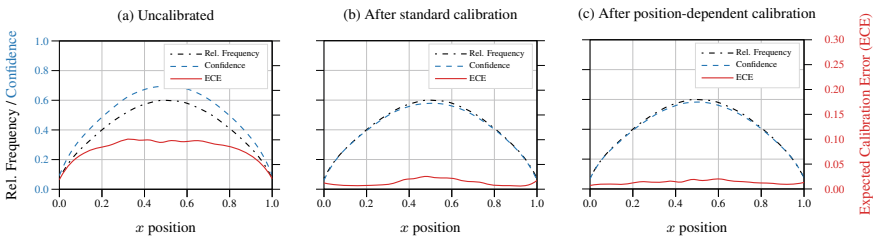

**Fig. 7** Reliability diagram (instance segmentation) for a `Mask R-CNN` class *pedestrian* inferred on MS COCO validation images measured by the relative $x$ position of each pixel within its bounding box. **a** Similar to the examinations for object detection, we observe an increasing calibration error toward the boundaries of the images. **b** Standard histogram binning significantly reduces the calibration error, **c** whereas position-dependent histogram binning is capable of a slightly further refinement of the position-dependent D-ECE

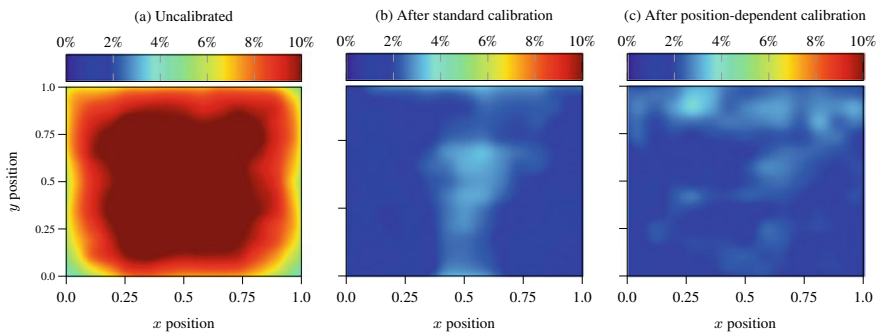

**Fig. 8** Reliability diagram (instance segmentation) for a `Mask R-CNN` class *pedestrian* inferred on MS COCO validation images. The diagram shows the D-ECE [%] that is measured using the relative $x$ and $y$ position of each pixel within its bounding box. Similar to the observations in Fig. 7, standard calibration already reduces the calibration error. However, we can observe a slightly increased calibration error toward the image's center. In contrast, position-dependent calibration error shows equal calibration performance across the whole image

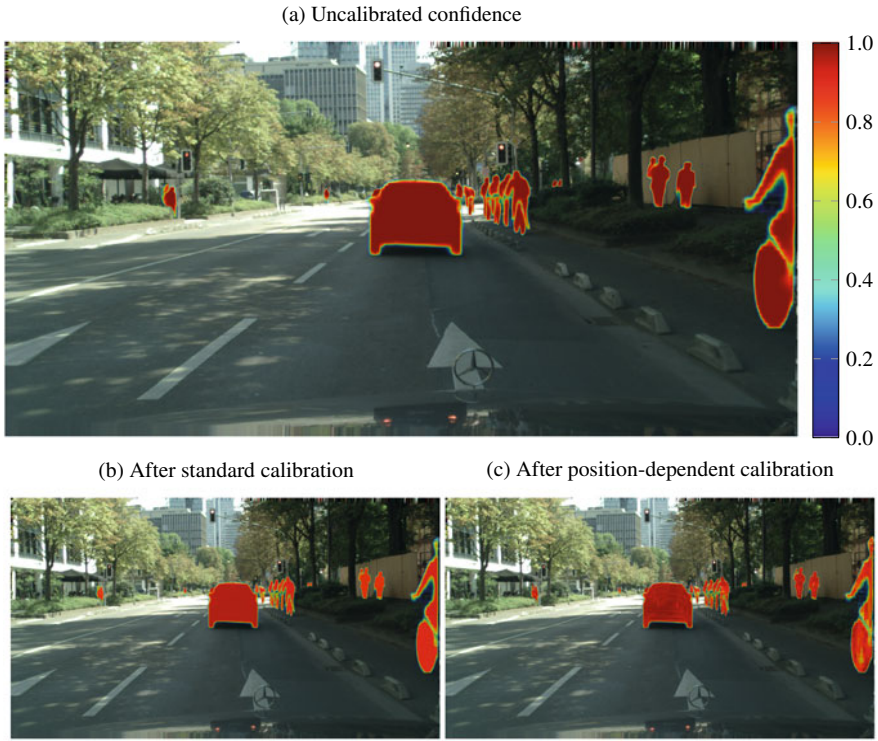

**Fig. 9** Qualitative example of a Cityscapes image with instance segmentation masks obtained by a Mask R-CNN. **a** Uncalibrated confidence of the predicted masks (mIoU = 0.289). **b** Instance masks are calibrated using standard histogram binning (mIoU = 0.343). **c** Mask confidences using a position-dependent confidence calibration (mIoU = 0.370). Unlike a simple scaling of confidences in **b**, the calibration in **c** offers a more fine-grained mapping that achieves a better fit to the true instance masks

## 5.3  Semantic Segmentation

For the evaluation of semantic segmentation, we use the same datasets, the same binning scheme, and the same features that have already been used for the instance segmentation in Sect. 5.2. For COCO mask inference, we use a pretrained Deeplabv2 [CPK+18] as well as a pretrained HRNet model [SXLW19, WSC+20, YCW20]. For Cityscapes, we also use a HRNet as well as a pretrained Deeplabv3+ model [CZP+18]. Similar to the instance segmentation, we also use the D-ECE, Brier score, NLL loss, and mIoU to compare the baseline calibration with a histogram binning model. As opposed to our previous experiments, we only use 15% of the provided samples to reduce computational complexity. The results for default miscalibration, standard calibration (subset: confidence only) and position-dependent calibration (subset: full) are shown in Table 5. Unlike instance segmentation calibration, the baseline model is already intrinsically well-calibrated, offering a very low calibration error. A qualitative example is shown in Fig. 10 to illustrate the effect of calibration for semantic segmentation. In this example, we can observe only minor differences between the uncalibrated mask and the masks either after standard calibration or after position-dependent calibration. This aspect is supported by the confidence reliability diagram shown in Fig. 11. Although we can observe an overconfidence, most samples are located in the low confident space with a low calibration error that also results in an overall low miscalibration score. Furthermore, our calibration methods are able to achieve even better calibration results, but mostly in the confidence-only calibration case. In addition, we also observe only a low correlation between position and calibration error (Figs. 12 and 13). One reason for the major difference between instance and semantic segmentation calibration may be the difference in model training. An instance segmentation model needs to infer an appropriate bounding box first to achieve qualitatively good results in mask inference. In contrast, a semantic segmentation model does not need to infer the position of objects within an image but is able to directly learn and improve mask quality. We also suspect an influence of the amount of available data points, since a semantic segmentation model is able to use each pixel within an image as a separate sample, whereas an instance segmentation model is only restricted to the pixels that are available within an estimated bounding box. Therefore, we conclude that semantic segmentation models do not require a post-hoc confidence calibration as they already offer a good calibration performance.

**Table 5** Calibration results for semantic segmentation. The best D-ECE scores are underlined *for each subset separately*, since it is not convenient to compare D-ECE scores with different subsets to each other [KKSH20]. Furthermore, the best Brier, NLL, and mIoU scores are highlighted in bold. These scores are only calculated using the confidence information and thus can be compared to each other. In contrast to object detection and instance segmentation evaluation, we observe a low baseline calibration error that is only slightly affected by confidence calibration

| Network | Dataset | Cal. method | Subset: confidence only | | | | Subset: full | | | |
|---|---|---|---|---|---|---|---|---|---|---|
| | | | D-ECE [%] | Brier | NLL | mIoU | D-ECE [%] | Brier | NLL | mIoU |
| Deeplabv3+ | CS | Baseline | 0.162 | **0.060** | **0.139** | **0.623** | 0.186 | **0.060** | **0.139** | **0.623** |
| | | HB | 0.081 | **0.060** | 0.170 | 0.619 | 0.199 | 0.062 | 0.189 | 0.589 |
| Deeplabv2 | COCO | Baseline | 0.094 | 0.458 | **1.173** | **0.933** | 0.149 | 0.458 | **1.173** | **0.933** |
| | | HB | 0.060 | **0.456** | 1.515 | **0.933** | 0.150 | 0.485 | 1.790 | 0.913 |
| HRNet | CS | Baseline | 0.067 | **0.057** | **0.115** | **0.629** | 0.149 | **0.057** | **0.115** | **0.629** |
| | | HB | 0.083 | **0.057** | 0.148 | 0.628 | 0.191 | 0.060 | 0.171 | 0.582 |
| | COCO | Baseline | 0.455 | 0.779 | 5.812 | **0.939** | 0.455 | 0.779 | 5.812 | **0.939** |
| | | HB | 0.062 | **0.563** | **2.261** | **0.939** | 0.138 | 0.571 | 2.372 | 0.931 |

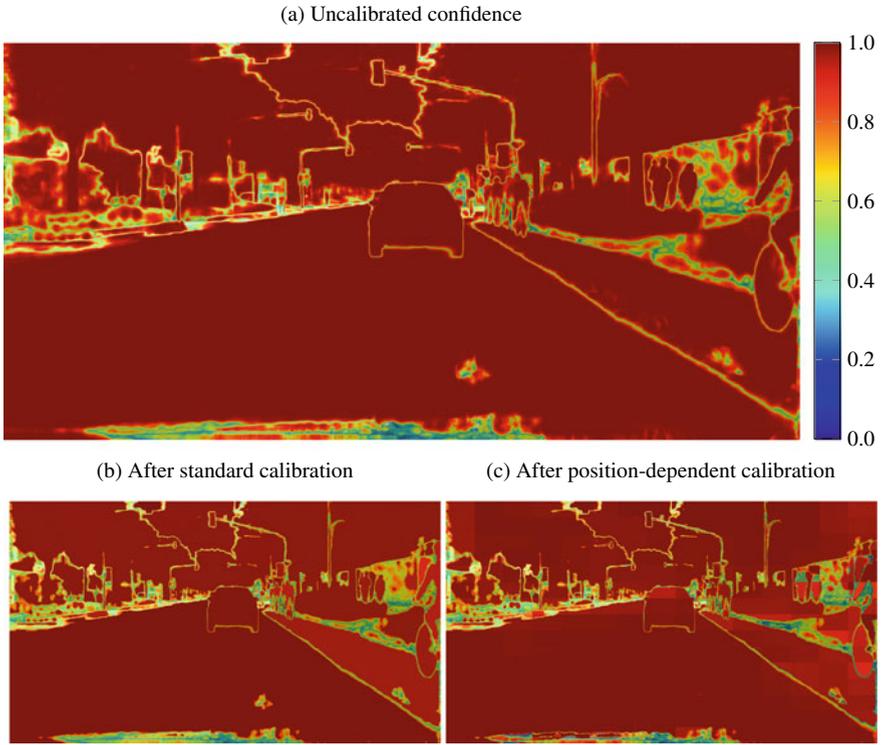

**Fig. 10** Qualitative example of a Cityscapes image with semantic segmentation masks obtained by a `Deeplabv3+` model. **a** Uncalibrated confidence of the predicted masks (mIoU = 0.576), **b** here, the pixel masks are calibrated using standard histogram binning (mIoU = 0.522). Furthermore, **c** mask confidences using a position-dependent confidence calibration (mIoU = 0.570). Although we observe some minor differences, we conclude that confidence calibration does not have a major influence to the calibration properties for semantic segmentation models

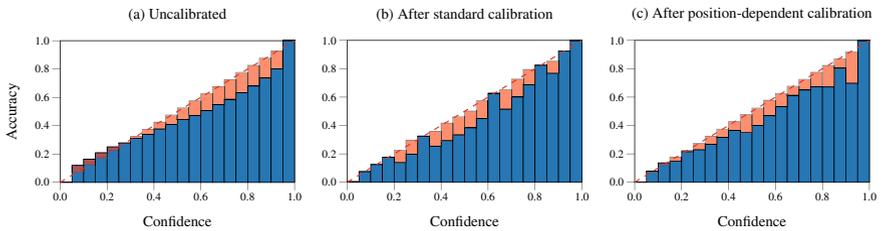

**Fig. 11** Reliability diagram (semantic segmentation) for a `Deeplabv3+` class *pedestrian* inferred on Cityscapes Frankfurt images measured using the confidence only. The blue bar denotes the accuracy of samples that fall into this bin, with the gap to perfect calibration (diagonal) highlighted in red. By default (**a**), the `Deeplabv3+` model already offers a good calibration performance that is only slightly affected either by standard calibration (**b**) or by position-dependent calibration (**c**)

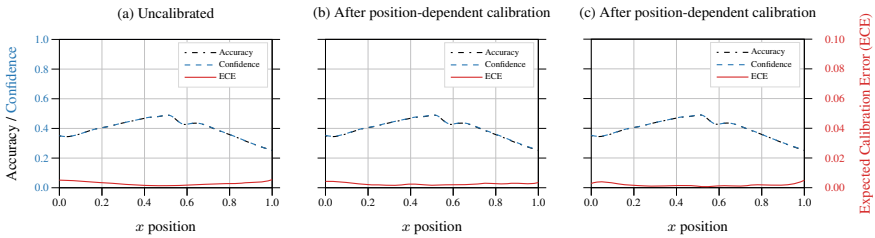

**Fig. 12** Reliability diagram (semantic segmentation) for a `Deeplabv3+` class *pedestrian* inferred on Cityscapes Frankfurt images measured by the relative *x* position of each pixel. Since the model has a low miscalibration by default (**a**), the effect of calibration in **b** or **c** is marginal

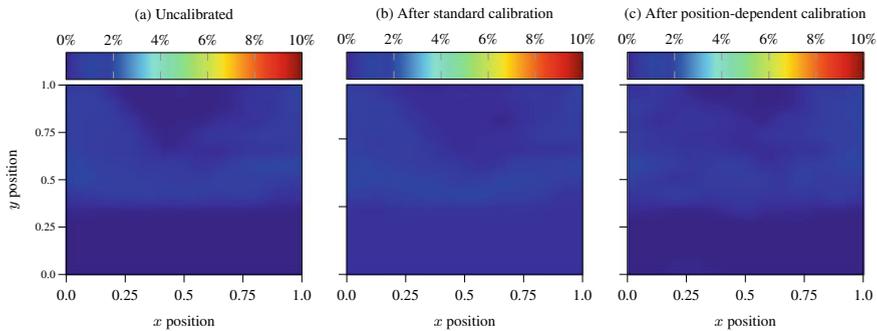

**Fig. 13** Reliability diagram (semantic segmentation) for a `Deeplabv3+` class *pedestrian* inferred on Cityscapes Frankfurt images. This diagram shows the D-ECE [%] that is measured using the relative *x* and *y* position of each pixel. Since the model has a low miscalibration by default (**a**), the effect of calibration in **b** or **c** is marginal

# Conclusions

Within the scope of confidence calibration, recent work has mainly focused on the task of classification calibration. In this chapter, we presented an analysis of confidence calibration for object detection calibration as well as, for instance, and semantic segmentation calibration. Firstly, we introduced definitions for confidence calibration within the scope of object detection, instance segmentation, and semantic segmentation. Secondly, we presented methods to measure and alleviate miscalibration of detection and segmentation networks. These methods are extensions of well-known calibration methods such as histogram binning [ZE01], Platt scaling [Pla99], and beta calibration [KSFF17]. We extend these methods so that they encompass additional calibration information such as position and shape. Finally, the experiments revealed that common object detection models as well as instance segmentation networks tend to miscalibration. In addition, we showed that auxiliary information such as estimated position or shape of a predicted object also have an influence to confidence calibration. However, we also found that semantic segmentation models are already intrinsically calibrated. Thus, the examined models do not require additional post-hoc calibration and already offer well-calibrated mask confidence scores. We argue that this difference between instance and semantic segmentation is a result of data quality and availability during training. This leads to the assumption that limited data availability is a direct cause for miscalibration during training and thus an effect of overfitting. Nevertheless, our proposed calibration framework is capable to calibrate object detection and instance segmentation models. In safety-critical applications, the confidence in the applied algorithms is paramount. The proposed calibration algorithms allow to detect situations of low confidence and thus perform the appropriate system reaction. Therefore, calibrated confidence values can be used as additional information especially in safety-critical applications.

**Acknowledgements** The research leading to these results is funded by the German Federal Ministry for Economic Affairs and Energy within the project "Methoden und Maßnahmen zur Absicherung von KI-basierten Wahrnehmungsfunktionen für das automatisierte Fahren (KI Absicherung)". The authors would like to thank the consortium for the successful cooperation.

# References

[COR+16] M. Cordts, M. Omran, S. Ramos, T. Rehfeld, M. Enzweiler, R. Benenson, U. Franke, S. Roth, B. Schiele, The cityscapes dataset for semantic urban scene understanding, in *Proceedings of the IEEE/CVF Conference on Computer Vision and Pattern Recognition (CVPR)* (Las Vegas, NV, USA, June 2016), pp. 3213–3223

[CPK+18] Liang-Chieh. Chen, George Papandreou, Iasonas Kokkinos, Kevin Murphy, Alan L. Yuille, DeepLab: semantic image segmentation with deep convolutional nets, atrous convolution, and fully connected CRFs. IEEE Trans. Pattern Anal. Mach. Intell. (TPAMI) **40**(4), 834–848 (2018)

[CZP+18] L.-C. Chen, Y. Zhu, G. Papandreou, F. Schroff, H. Adam, Encoder-decoder with atrous separable convolution for semantic image segmentation, in *Proceedings of the European Conference on Computer Vision (ECCV)* (Munich, Germany, September 2018), pp. 833–851

[DLXS20] Y. Ding, J. Liu, J. Xiong, Y. Shi, Revisiting the evaluation of uncertainty estimation and its application to explore model complexity-uncertainty trade-off, in *Proceedings of the IEEE/CVF Conference on Computer Vision and Pattern Recognition (CVPR) Workshops* virtual conference (June 2020), pp. 22–31

[GPSW17] C. Guo, G. Pleiss, Y. Sun, K.Q. Weinberger, On calibration of modern neural networks, in *Proceedings of the International Conference on Machine Learning (ICML)* (Sydney, NSW, Australia, August 2017), pp. 1321–1330

[HGDG17] K. He, G. Gkioxari, P. Dollár, R. Girshick, Mask R-CNN, in *Proceedings of the IEEE International Conference on Computer Vision (ICCV)* (Venice, Italy, October 2017), pp. 2980–2988

[JJY+19] B. Ji, H. Jung, J. Yoon, K. Kim, Y. Shin, Bin-wise temperature scaling (BTS): improvement in confidence calibration performance through simple scaling techniques, in *Proceedings of the IEEE International Conference on Computer Vision (ICCV) Workshops* (Seoul, Korea, October 2019), pp. 4190–4196

[KG20] D. Karimi, A. Gholipour, Improving Calibration and Out-of-Distribution Detection in Medical Image Segmentation With Convolutional Neural Networks (May 2020), pp. 1–12. arXiv:2004.06569

[KKSH20] F. Küppers, J. Kronenberger, A. Shantia, A. Haselhoff, Multivariate confidence calibration for object detection, in *Proceedings of the IEEE/CVF Conference on Computer Vision and Pattern Recognition (CVPR) Workshops*, virtual conference (June 2020), pp. 1322–1330

[KLM19] A. Kumar, P.S. Liang, T. Ma, Verified uncertainty calibration, in *Proceedings of the Conference on Neural Information Processing Systems (NIPS/NeurIPS)* (Vancouver, BC, Canada, December 2019), pp. 3787–3798

[KPNK+19] M. Kull, M. Perello Nieto, M. Kängsepp, T. Silva Filho, H. Song, P. Flach, Beyond temperature scaling: obtaining well-calibrated multi-class probabilities with Dirichlet calibration, in *Proceedings of the Conference on Neural Information Processing Systems (NIPS/NeurIPS)* (Vancouver, BC, Canada, December 2019), pp. 12316–12326

[KSFF17] M. Kull, T. Silva Filho, P. Flach, Beta calibration: a well-founded and easily implemented improvement on logistic calibration for binary classifiers, in *Proceedings of the International Conference on Artificial Intelligence and Statistics (AISTATS)* (Fort Lauderdale, FL, USA, May 2017), pp. 623–631

[KWHG20] A. Kirillov, Y. Wu, K. He, R. Girshick, PointRend: image segmentation as rendering, in *Proceedings of the IEEE/CVF Conference on Computer Vision and Pattern Recognition (CVPR)*, virtual conference (June 2020), pp. 9799–9808

[LGG+17] T.-Y. Lin, P. Goyal, R. Girshick, K. He, P. Dollár, Focal loss for dense object detection, in: *Proceedings of the IEEE International Conference on Computer Vision (ICCV)* (Venice, Italy, October 2017), pp. 2980–2988

[LMB+14] T.-Y. Lin, M. Maire, S. Belongie, J. Hays, P. Perona, D. Ramanan, P. Dollár, C. Lawrence Zitnick, Microsoft COCO: common objects in context, in *Proceedings of the European Conference on Computer Vision (ECCV)* (Zurich, Switzerland, September 2014), pp. 740–755

[LN82] David L. Libby, Melvin R. Novick, Multivariate generalized beta distributions with applications to utility assessment. J. Educ. Stat. **7**(4), 271–294 (1982)

[MWT+20] Alireza Mehrtash, William M. Wells, Clare M. Tempany, Purang Abolmaesumi, Tina Kapur, Confidence calibration and predictive uncertainty estimation for deep medical image segmentation. IEEE Trans. Med. Imaging **39**(12), 3868–3878 (2020)

[NC16] M. Naeini, G. Cooper, Binary classifier calibration using an ensemble of near isotonic regression models, in *Proceedings of the IEEE International Conference on Data Mining (ICDM)* (Barcelona, Spain, December 2016), pp. 360–369

[NCH15] M. Pakdaman Naeini, G. Cooper, M. Hauskrecht, Obtaining well calibrated probabilities using Bayesian binning, in *Proceedings of the AAAI Conference on Artificial Intelligence* (Austin, TX, USA, January 2015), pp. 2901–2907

[NDZ+19] J. Nixon, M.W. Dusenberry, L. Zhang, G. Jerfel, D. Tran, Measuring calibration in deep learning, in *Proceedings of the IEEE/CVF Conference on Computer Vision and Pattern Recognition (CVPR) Workshops* (Long Beach, CA, USA, June 2019), pp. 38–41

[Pla99] J. Platt, Probabilistic outputs for support vector machines and comparisons to regularized likelihood methods, in *Advances in Large Margin Classifiers*. ed. by A.J. Smola, P. Bartlett, B. Schölkopf, D. Schuurmans (MIT Press, 1999), pp. 61–74

[RHGS15] S. Ren, K. He, R. Girshick, J. Sun, Faster R-CNN: towards real-time object detection with region proposal networks, in *Proceedings of the Conference on Neural Information Processing Systems (NIPS/NeurIPS)* (Montréal, QC, Canada, December 2015), pp. 91–99

[SKR+21] F. Schwaiger, M. Henne Fabian Küppers, F. Schmoeller Roza, K. Roscher, A. Haselhoff, From black-box to white-box: examining confidence calibration under different conditions, in *Proceedings of the Workshop on Artificial Intelligence Safety (SafeAI)*, virtual conference (February 2021), pp. 1–8

[SXLW19] K. Sun, B. Xiao, D. Liu, J. Wang, Deep high-resolution representation learning for human pose estimation, in *Proceedings of the IEEE/CVF Conference on Computer Vision and Pattern Recognition (CVPR)* (Long Beach, CA, USA, June 2019), pp. 5693–5703

[WKM+19] Y. Wu, A. Kirillov, F. Massa, W.-Y. Lo, R. Girshick, Detectron2 (2019). [Online; assessed 2021-11-18]
[WLK+20] T. Wang, Y. Li, B. Kang, J. Li, J. Liew, S. Tang, S. Hoi, J. Feng, The devil is in classification: a simple framework for long-tail instance segmentation, in *Proceedings of the European Conference on Computer Vision (ECCV)*, virtual conference (August 2020), pp. 728–744
[WSC+20] Jingdong Wang, Ke. Sun, Tianheng Cheng, Borui Jiang, Chaorui Deng, Yang Zhao, Dong Liu, Mu. Yadong, Mingkui Tan, Xinggang Wang et al., Deep high-resolution representation learning for visual recognition. IEEE Trans. Pattern Anal. Mach. Intell. (TPAMI) **43**(10), 3349–3364 (2020)
[YCW20] Y. Yuan, X. Chen, J. Wang, Object-contextual representations for semantic segmentation, in *Proceedings of the European Conference on Computer Vision (ECCV)*, virtual conference (August 2020), pp. 173–190
[ZE01] B. Zadrozny, C. Elkan, Obtaining calibrated probability estimates from decision trees and Naive Bayesian classifiers, in *Proceedings of the International Conference on Machine Learning (ICML)* (Williamstown, MA, USA, June 2001), pp. 609–616
[ZE02] B. Zadrozny, C. Elkan, Transforming classifier scores into accurate multiclass probability estimates, in *Proceedings of the ACM SIGKDD International Conference on Knowledge Discovery and Data Mining (KDD)* (Edmonton, AB, Canada, July 2002), pp. 694–699